\RequirePackage{fix-cm}
\documentclass[number,preprint,review,1p,times]{elsarticle}

\usepackage{mathalfa}
\usepackage{hyperref}

\usepackage{graphicx}
\usepackage{tabularx}

\usepackage{pgf-pie}
\usepackage{subfiles}
\usepackage{graphicx}
\usepackage[utf8]{inputenc}
\usepackage{comment}
\usepackage[english]{babel}
\usepackage{textcomp}
\usepackage{adjustbox}
\usepackage{indentfirst}
\usepackage{mathtools}
\usepackage[usestackEOL]{stackengine}
\usepackage{ltxtable}

\usepackage{amsmath}
\usepackage{amsfonts}

\usepackage{algorithm}
\usepackage{algorithmicx}
\usepackage{algpseudocode}

\usepackage{textcomp}
\usepackage{mathpazo}
\usepackage{tgpagella}
\usepackage{color,soul}
\usepackage{dirtytalk}
\usepackage{tcolorbox}
\usepackage{bm}
\usepackage{graphicx,subcaption,chngpage,calc}
 \usepackage[math]{cellspace}
  \usepackage{ltablex}
\usepackage{xtab, lipsum, array}
\usepackage{caption}
\usepackage{multirow}
\usepackage{tabularx,ragged2e,booktabs,caption}
\newcolumntype{L}{>{\raggedright\arraybackslash}X}

\stackMath

\DeclareMathSizes{12}{17.28}{9}{7}
\usepackage{color}
\usepackage{multirow, makecell}
\usepackage{ltablex}

\usepackage{url}

\usepackage{tabularx}
\usepackage{pgfplots}
\pgfplotsset{compat=1.14}
\usepackage{pgfplotstable}
\usepackage{array, booktabs, makecell}
\usepackage{color}
\usepackage{csquotes}
\usepackage{url}
\usepackage{comment}
\usepackage{tikz}
\usepackage{balance}
\usepackage{listings}
\usepackage{breqn}
\usepackage{booktabs} 
\usepackage{soul} 
\usetikzlibrary{fit} 
\usetikzlibrary{positioning}
\usetikzlibrary{arrows}
\usetikzlibrary{shapes.multipart}
\usepackage{multirow} 
\usepackage{multicol}
\usepackage{varwidth} 

\usepackage{multirow} 
\usepackage{multicol}
\usepackage{varwidth} 
\usepackage{makecell}
\usepackage{adjustbox}

\usepackage{tikz}
\usetikzlibrary{shapes,arrows,positioning,fit,backgrounds}
\usetikzlibrary{arrows.meta, calc}

\pgfdeclarelayer{background}
\pgfsetlayers{background,main}

\definecolor{Gray}{gray}{0.80} 

\definecolor{bblue}{HTML}{4F81BD}
\definecolor{rred}{HTML}{C0504D}
\definecolor{ggreen}{HTML}{9BBB59}
\definecolor{ppurple}{HTML}{9F4C7C}

\newcommand{\moxe}{MoxE}
\newcommand{\mlstm}{mLSTM}
\newcommand{\slstm}{sLSTM}

\newcommand{\ffn}{FFN}
\newcommand{\xlstmmoe}{xLSTM-MoE}
\newcommand{\mlstmmoe}{mLSTM-MoxE}
\newcommand{\slstmmoe}{sLSTM-MoxE}
\newcommand{\nogrouploss}{MoxE-No-Group-Loss}
\newcommand{\nobias}{MoxE-Standard}

\begin{document}
\begin{frontmatter}

\title{MoxE: Mixture of xLSTM Experts with Entropy-Aware Routing for Efficient Language Modeling}

\author[First]{Abdoul Majid O. Thiombiano}
\ead{abdoulmajid.ousseinithiombiano@fsm.rnu.tn}

\author[First,Second]{Brahim Hnich}
\ead{brahim.hnich@fsm.rnu.tn}

\author[Third]{Ali Ben Mrad}
\ead{a.benmrad@qu.edu.sa}

\author[Fourth]{Mohamed Wiem Mkaouer}
\ead{mmkaouer@umich.edu}

\address[First]{FSM, University of Monastir, Monastir, 5000 Tunisia}
\address[Second]{CES Lab, ENIS, University of Sfax, Sfax, 3038 Tunisia}
\address[Third]{Department of Computer Science, College of Computer, Qassim University, Buraydah, Saudi Arabia}
\address[Fourth]{University of Michigan-Flint, MI, USA}



\begin{abstract}
This paper introduces \textbf{MoxE}, a novel architecture that synergistically combines the Extended Long Short-Term Memory (xLSTM) with the Mixture of Experts (MoE) framework to address critical scalability and efficiency challenges in large language models (LLMs). The proposed method effectively leverages xLSTM's innovative memory structures while strategically introducing sparsity through MoE to substantially reduce computational overhead. At the heart of our approach is a novel entropy-based routing mechanism, designed to dynamically route tokens to specialized experts, thereby ensuring efficient and balanced resource utilization. This entropy awareness enables the architecture to effectively manage both rare and common tokens, with mLSTM blocks being favored to handle rare tokens. To further enhance generalization, we introduce a suite of auxiliary losses, including entropy-based and group-wise balancing losses, ensuring robust performance and efficient training. Theoretical analysis and empirical evaluations rigorously demonstrate that MoxE achieves \textbf{significant efficiency gains and enhanced effectiveness} compared to existing approaches, marking a notable advancement in scalable LLM architectures.
\end{abstract}

\end{frontmatter}

\section{Introduction}
The NLP space is predominantly dominated by attention-based models \cite{waswani2017attention}, which have demonstrated remarkable capabilities across various language tasks. However, the quadratic complexity $O(n^2)$ of the attention mechanism (where $n$ is the sequence length) makes it computationally expensive to train and deploy large models, particularly for long sequences. This inherent limitation poses significant challenges for scalability and efficiency in real-world applications.

One highly effective technique widely adopted to mitigate these challenges in training and deploying such massive models is the Mixture of Experts (MoE) framework \cite{jacobs1991adaptive, shazeer2017outrageouslylargeneuralnetworks}. By design, in a MoE architecture, at inference time, the model intelligently utilizes only a sparse subset of its total parameters to process each input, leading to a dramatic reduction in the computational requirements at runtime and enabling more efficient scaling. The sparse MoE approach has been successfully applied to various models, demonstrating significant improvements in efficiency while maintaining or even enhancing performance \cite{fedus2022switch}.

Traditional Long Short-Term Memory (LSTM) networks, while demonstrably powerful in sequence modeling, inherently struggle with effectively managing long-term dependencies and achieving efficient associative recall, particularly when dealing with extended sequences. The Extended Long Short-Term Memory (xLSTM) architecture \cite{beck2025xlstm} directly addresses these fundamental limitations by introducing novel memory structures and optimized computation approaches within the LSTM unit itself. xLSTM offers improved performance for individual recurrent units and further demonstrates efficient memory usage.

Building upon these advancements, in this paper, we propose \textbf{MoxE}, a novel architecture that thoughtfully combines the inherent strengths of the xLSTM unit with the sparsity-inducing and efficiency-enhancing properties of the Mixture of Experts framework. This synergistic combination allows us to leverage the improved memory and computational efficiency of xLSTM at the unit level, while simultaneously addressing the scalability challenges of deploying very large models through MoE. Furthermore, we introduce the concept of entropy into our routing mechanism to effectively handle difficult-to-predict tokens. By explicitly teaching the model to preferentially utilize mLSTM experts for high-entropy (rare and complex) tokens, we aim to optimize resource allocation and further enhance the model's performance and efficiency.

\section{Background}
\label{sec:related}

\subsection{Extended Long Short-Term Memory (xLSTM)}
The xLSTM architecture \cite{beck2025xlstm} extends traditional LSTMs by introducing two novel computational units that address the limitations of standard recurrent models. The first unit, sLSTM, enhances the traditional LSTM layer with a novel memory-mixing technique and 
the second unit, mLSTM, expands the LSTM memory to a $d \times d$ matrix and is designed to be parallelizable like a transformer, making it ideal for tasks requiring high recall capabilities.

These innovations provide xLSTM with a linear complexity $O(n)$ during training and constant complexity $O(1)$ at inference time, offering a significant advantage over the quadratic complexity of traditional transformer models. This efficiency makes xLSTM particularly well-suited for processing long sequences, where attention-based models become computationally prohibitive.

\subsection{Mixture of Experts (MoE)}
The Mixture of Experts (MoE) framework, introduced by Jacobs et al.~\cite{jacobs1991adaptive}, is a neural architecture that combines multiple expert networks to process inputs adaptively. Each expert $E_i$ is typically a feed-forward network, and a gating network $G(\cdot)$ dynamically assigns input tokens to experts based on their relevance. The final output $y$ of a model's $l$-th MoE layer is computed as a weighted sum of expert outputs, where the gating network determines the weights:

\begin{equation}
y^{(l)} = \sum_{i=1}^{N} G^{(l)}(x)_i \cdot E^{(l)}_i(x)
\end{equation}

Here, $x$ is the input vector, $N$ is the total number of experts, and $G(x)_i$ represents the gating network's probability of selecting expert $i$. The gating network often uses a softmax function to normalize probabilities:

\begin{equation}
G(x)_i = \frac{\exp(w_i^T x + b_i)}{\sum_{j=1}^{N} \exp(w_j^T x + b_j)}
\end{equation}

where $w_i$ and $b_i$ are learnable parameters for expert $i$. This allows the model to allocate computational resources to experts most suited for the input, improving specialization and efficiency. Recent advancements, such as sparsely-gated MoE layers~\cite{shazeer2017outrageouslylargeneuralnetworks}, enable scaling to trillion-parameter models by leveraging conditional computation, where only a subset of experts are activated per input~\cite{fedus2022switch}.

\subsection{Sparsely-Gated MoE in Large Language Models}

Sparsely-gated MoE architectures, such as the Switch Transformer~\cite{fedus2022switch}, extend the MoE framework by activating only a subset of experts for each input token. This reduces computational costs while maintaining model capacity. The key components include:

\begin{itemize}
    \item \textbf{Router Selection}: A router $R$ selects the top-$K$ experts for each token using a sparse gating mechanism. The router logits $r_i = w_i^T h + b_i$ are computed for each expert, and the top-$K$ experts are chosen via:
    
    \begin{equation}
    \text{TopK}(R(h)) = \{ i \mid r_i \in \text{top-}K \text{ values of } r_1, r_2, \dots, r_N \}
    \end{equation}
    
    The router logits are then normalized using softmax. This ensures that only the top-$K$ experts contribute to the output~\cite{fedus2022switch}.
    
    \item \textbf{Efficient Computation}: By limiting expert activation to a subset (e.g., $K=2$ or $K=8$), the computational load scales linearly with $K$ rather than $N$. For example, in the Switch Transformer \cite{fedus2022switch}, a top-1 routing scheme was used to scale models to 1.6 trillion parameters while maintaining constant FLOPs.
    
    \item \textbf{Load Balancing}: Techniques like expert capacity thresholds~\cite{lepikhin2020gshardscalinggiantmodels} prevent overloading specific experts by capping the number of tokens assigned to each expert. If an expert's capacity is exceeded, excess tokens are dropped or redistributed.

    \item \textbf{Expert Capacity}:
    Each expert processes at most $C \times \frac{T}{N}$ tokens, where $C$ is a hyperparameter and $T$ is the total number of tokens~\cite{lepikhin2020gshardscalinggiantmodels}.
\end{itemize}

\subsection{Recent Advancements in MoE Routing}

Recent research has identified inherent uncertainty in MoE router modules, which can sometimes lead to suboptimal expert selection. Huang et al. \cite{huang2024hardertasksneedexperts} demonstrated that leveraging this uncertainty can actually enhance model performance by dynamically allocating more experts to more difficult tasks. Their work showed that harder language modeling tasks benefit from increased expert allocation, suggesting that adaptive routing strategies based on task difficulty can significantly improve MoE model performance.

On the other hand, works such as GW-MoE \cite{wu2024gwmoeresolvinguncertaintymoe} have addressed this challenge by improving model performance during fine-tuning of MoE models
by using the router's uncertainty as an indicator signal to forward the token to all experts instead of only a subset of them, given that the inherent uncertainty in router modules sometimes leads to uniform expert selection, which can be inefficient.

Additionally, recent explorations into integrating recurrent models with the MoE framework, particularly Mamba-based approaches \cite{pióro2024moemambaefficientselectivestate, lieber2024jambahybridtransformermambalanguage}, have shown promising results. These approaches demonstrate that recurrent architectures can be effectively combined with MoE techniques to create efficient and powerful language models.

\section{MoxE Architecture}
\label{sec:model}

\subsection{Overview}

\begin{figure}[H]
    \centering
    \resizebox{\linewidth}{!}{\includegraphics{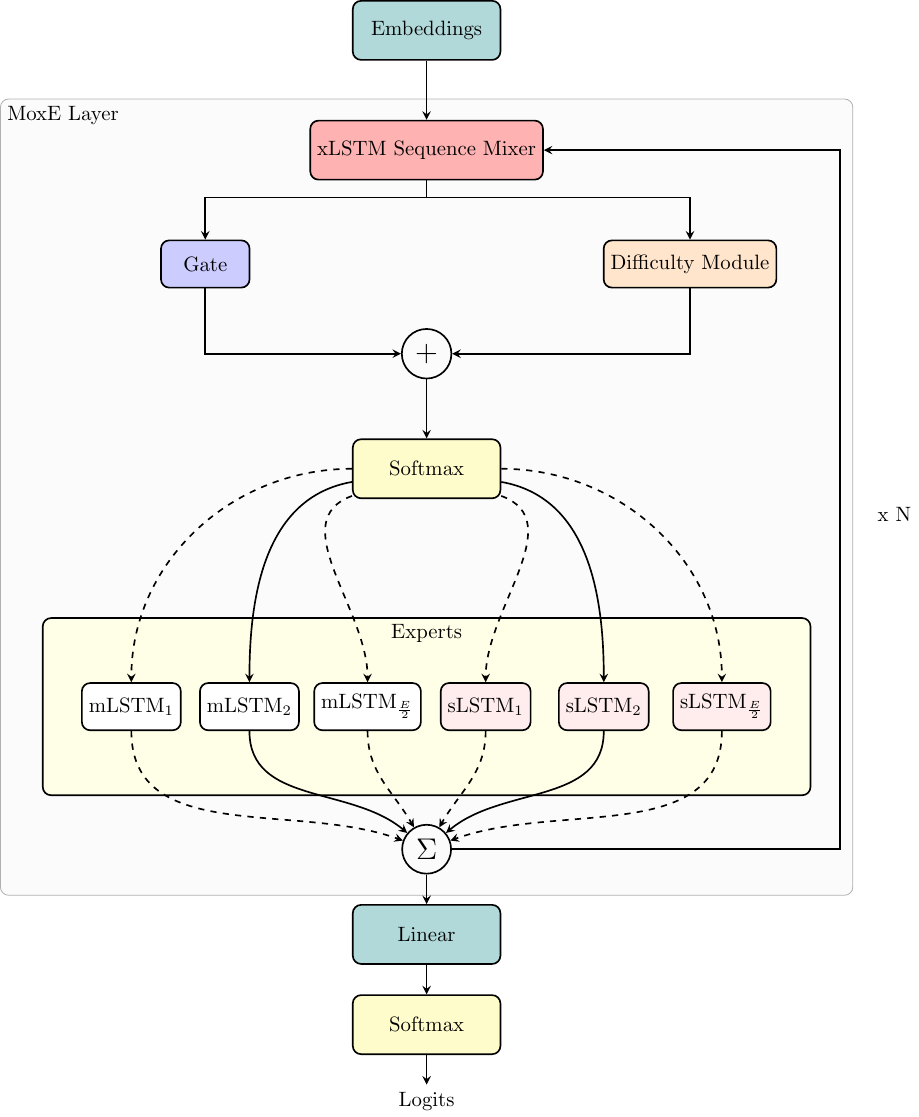}}
    \caption{The MoxE architecture with an xLSTM sequence mixer (composed with one sLSTM unit and one mLSTM unit).}
    \label{fig:moxe_architecture}
\end{figure}

\begin{figure}
    \centering
    \scalebox{0.55}{\includegraphics{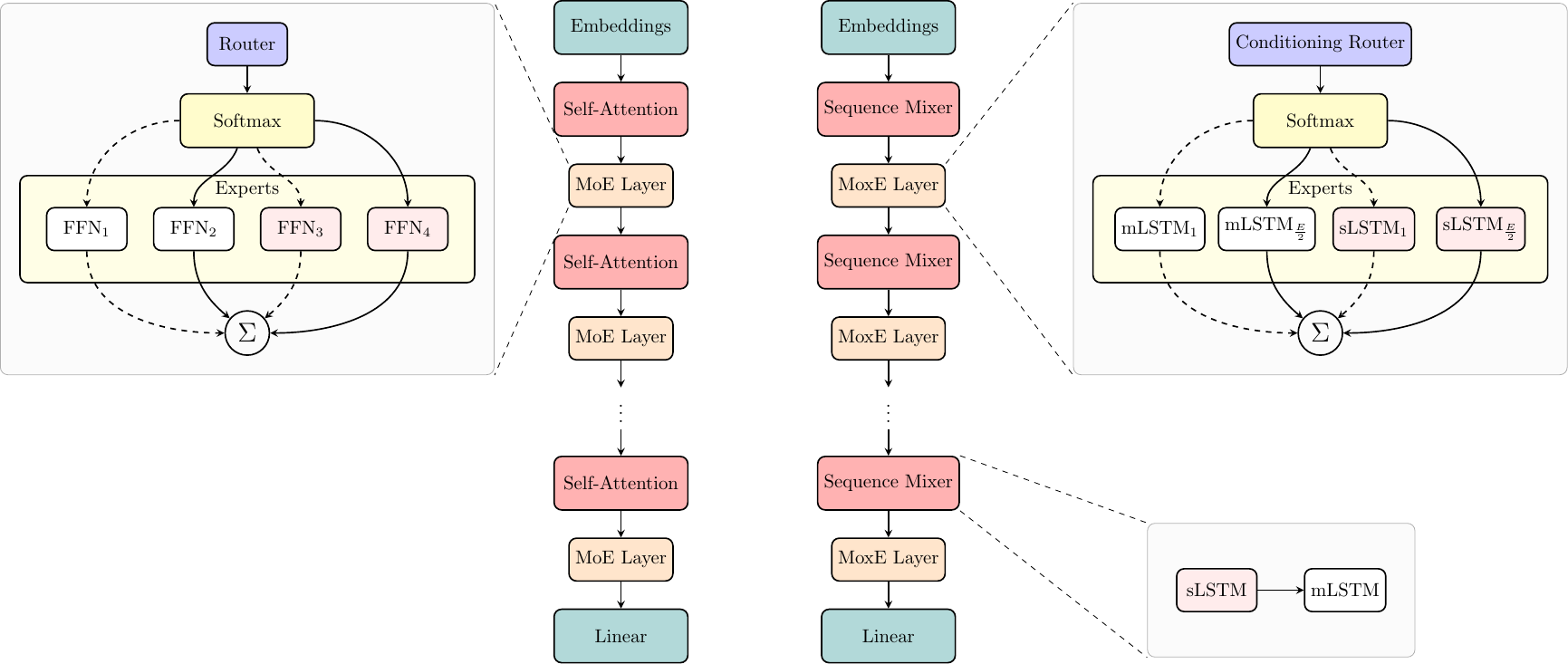}}
    \caption{A side-by-side comparison of an attention-based MoE model (on the left) and a MoxE model (on the right).}
    \label{fig:architectures_comparison}
\end{figure}

MoxE is a novel architecture (Figure~\ref{fig:moxe_architecture}) that leverages the efficiency of xLSTM units within an MoE framework. The key innovation lies in our entropy-aware routing mechanism that dynamically directs tokens to the appropriate expert type based on their complexity. Our architecture makes two fundamental changes to the traditional Transformer-based MoE framework:

\begin{itemize}
    \item We utilize the router's inherent uncertainty to modulate the expert selection process, leveraging the unique properties of the two computational units introduced by xLSTM.
    \item We utilize an xLSTM-based sequence mixer instead of the attention block and replace feed-forward experts with mLSTM and sLSTM blocks (Figure~\ref{fig:architectures_comparison}), creating a fully recurrent MoE model that benefits from both the efficiency of the sparse computation provided by the modern MoE framework and the powerful sequential modeling capabilities of xLSTM.
\end{itemize}

\subsection{Difficulty Assessment and Router Biasing}
A key innovation in our approach is the difficulty module $D$, which computes a per-token scalar value $d_t$ that represents the router's uncertainty when selecting the most suitable experts to handle the current token. Given an input sequence $x = \{x_1, x_2, \ldots, x_S\}$, where $x_t \in \mathbb{R}^d$, $S$ is the sequence length, and $d$ is the embedding size, the model processes tokens through a series of MoxE layers. For each token $x_t$, the model first computes a hidden state:

\begin{equation}
h_t = \text{Embedding}(x_t)
\end{equation}

In our case, we define $D$ as a linear projection such that: $D: \mathbb{R}^d \rightarrow [0,1], h_t \mapsto d_t \, \text{where} \, d_t \in [0,1]$ and the difficulty $d_t$ for token $x_t$ is then computed as:

\begin{equation} \label{eq:difficulty_score}
d_t = D(h_t) = \sigma(w_{D}^Th_t + b_D) \in [0, 1]
\end{equation}

with $\sigma(\cdot)$ being the sigmoid function. This difficulty score is used to bias the router's decision, encouraging it to route difficult-to-predict tokens towards mLSTM-based experts that have a greater recall capacity due to their matrix memory.

The router computes raw logits for expert selection $\tilde{z_t} = G(h_t) \in \mathbb{R}^E$ where $E$ is the total number of experts, with $n_{mLSTM} = n_{sLSTM} = \frac{E}{2}$. Let $\gamma > 0$ be a hyperparameter that scales the influence of the difficulty score on routing decisions, we define a modulation bias $\delta_{t,i}$ per token for each expert based on the token's difficulty score $d_t$ as:

\begin{align} \label{eq:modulation_bias}
    \delta_{t,i} &= 
        \begin{cases}
        \gamma d_t, & \text{if } i \in \text{mLSTM experts} \\
        -\gamma d_t, & \text{if } i \in \text{sLSTM experts}
        \end{cases}
\end{align}

This bias is then added to the raw logits $\tilde{z_t}$ to produce the adjusted logits:

\begin{equation}
z_t = \tilde{z_t} + \delta_t
\end{equation}

The routing probabilities are computed using softmax:

\begin{equation}
p_{t,i} = \frac{\exp(z_{t,i})}{\sum_{j=1}^{E} \exp(z_{t,j})}
\end{equation}

The final output is computed as a weighted sum of top-$k$ experts' output:

\begin{equation}
y_t = \sum_{k=1}^{topK} p_{t,k} E_k(h_t)
\end{equation}

where $E_i$ represents either an mLSTM or sLSTM expert, depending on the expert index.

\section{Loss Functions and Balancing Strategies}
\label{sec:losses}

Training MoxE involves a combination of task-specific losses and auxiliary losses that ensure efficient training and balanced expert utilization. These auxiliary losses are crucial for maintaining stability and encouraging the desired routing behavior.

\subsection{Auxiliary Difficulty Loss}

We introduce an auxiliary difficulty loss that encourages the difficulty prediction to align with the token's normalized entropy:

\begin{equation}
\mathcal{L}_{d} = \frac{1}{B \times S} \sum_{t,b} \left(d_{t,b} - \tilde{H}_{t,b}\right)^2
\end{equation}

where $\tilde{H}_{t,b}$ is the normalized entropy computed from the \textit{unbiased} routing probabilities $p*$:

\begin{equation}
\tilde{H}_{t,b} = \frac{-\sum_{i=1}^{E} \tilde{p}_{t,b,i} \log \tilde{p}_{t,b,i}}{\log E}
\end{equation}

This loss ensures that tokens with high entropy (indicating uncertainty in prediction) are appropriately routed to the more capable mLSTM experts.

\subsection{Group-Wise Auxiliary Loss}

To maintain balance between the two expert groups (mLSTM and sLSTM), we introduce a group-wise auxiliary loss:

\begin{equation}
p_{m} = \frac{1}{B \times S} \sum_{t,b} \sum_{i \in \text{\mlstm}} p_{t,b}(i)
\end{equation}

\begin{equation}
p_{s} = \frac{1}{B \times S} \sum_{t,b} \sum_{i \in \text{\slstm}} p_{t,b}(i)
\end{equation}

\begin{align}
\mathcal{L}_{group} &= \text{KL}([p_m, p_s] || [0.5, 0.5]) \\
&= p_m \log \frac{p_m}{0.5} + p_s \log \frac{p_s}{0.5} \\
&= p_m \log(p_m) - p_m \log(0.5) + p_s \log(p_s) - p_s \log(0.5) \\
&= p_m \log(p_m) + p_m \log(2) + p_s \log(p_s) + p_s \log(2) \\
&= p_m \log(2 \cdot p_m) + p_s \log(2 \cdot p_s)
\end{align}

This loss encourages a balanced utilization of both expert types across the batch, preventing the model from consistently favoring one expert type over the other.

\subsection{Router Z-Loss}

To stabilize router logits and prevent extreme values, we incorporate a router Z-loss \cite{fedus2022switch}:

\begin{equation}
\mathcal{L}_{z} = \frac{1}{B \times S} \sum_{t,b} \left(\log \sum_{i=1}^{E} \exp(\tilde{z}_{t,b,i})\right)^2
\end{equation}

This loss penalizes large logit values, which can lead to overly confident routing decisions and potentially inefficient expert utilization.

\subsection{Load Balancing Auxiliary Loss}

To prevent certain experts from being overloaded or underutilized, we employ a load-balancing auxiliary loss \cite{fedus2022switch}:

\begin{equation}
\mathcal{L}_{aux} = E_s \sum_{i=1}^{E} \frac{p_i f_i}{p_i}
\end{equation}

where:
\begin{itemize}
    \item $L_{aux}$ is the auxiliary loss that promotes balanced expert usage.
    \item $E$ is the total number of experts in the MoE model.
    \item $p_i$ is the expected fraction of tokens routed to the $i$-th expert, computed from the router's softmax output.
    \item $f_i$ is the actual fraction of tokens processed by the $i$-th expert.
    \item $E_s$ is a scaling factor used to stabilize the loss.
\end{itemize}

\subsection{Total Loss}

The final training objective is a weighted combination of the task-specific loss (e.g., language modeling loss) and the auxiliary losses:

\begin{equation}
L_{total} = \mathcal{L}_{task} + \lambda_d \mathcal{L}_{d} + \lambda_{group} \mathcal{L}_{group} + \lambda_z L_z + \lambda_{aux} \mathcal{L}_{aux}
\end{equation}

where $\lambda_d$, $\lambda_{group}$, $\lambda_z$, and $\lambda_{aux}$ are hyperparameters that control the contribution of each auxiliary loss to the total loss.

\section{Theoretical Analysis}
\label{sec:analysis}

\subsection{Computational Efficiency}

The computational efficiency of MoxE stems from two key factors: the linear complexity of xLSTM and the sparsity introduced by the MoE framework. The xLSTM architecture has a time complexity of $O(n)$ during training and $O(1)$ during inference, where $n$ is the sequence length. This is already a significant improvement over the $O(n^2)$ complexity of transformer-based models.

With the MoE framework, we further reduce computational costs by activating only a subset of experts for each token. For a model with $E$ experts and $k$ active experts per token (where $k \ll E$), the computational cost is effectively reduced by a factor of $\frac{k}{E}$ compared to a dense model with equivalent capacity. The overall complexity of MoxE can be expressed as:

\begin{equation}
\text{Cost}_{MoxE} = O(n) \times \frac{k}{E} = O\left(\frac{nk}{E}\right)
\end{equation}

This makes MoxE particularly efficient for large-scale applications, allowing it to scale to larger model sizes without proportionally increasing computational requirements.

\subsection{Entropy-Based Routing Analysis}

The effectiveness of our entropy-based routing mechanism can be analyzed in terms of its ability to match token difficulty with expert capability. For a token with difficulty $d_t$, the probability of routing to an mLSTM expert versus an sLSTM expert is influenced by the bias term $\delta_t$.

The total probability of routing a token $t$ to any expert in the mLSTM group is the sum of the probabilities for individual mLSTM experts:

\begin{equation}
    P(\text{\mlstm}|d_t) = \sum_{i \in \text{\mlstm}} p_{t,i}
\end{equation}
Similarly, the total probability of routing to the sLSTM group is:

\begin{equation}
    P(\text{\slstm}|d_t) = \sum_{k \in \text{\slstm}} p_{t,k}
\end{equation}

\subsection*{Derivation of the Probability Ratio}

We want to find the ratio $\frac{P(\text{\mlstm}|d_t)}{P(\text{\slstm}|d_t)}$.

First, let's express $P(\text{\mlstm}|d_t)$ using the definitions:
\begin{align}
P(\text{\mlstm}|d_t) &= \sum_{i \in \text{\mlstm}} p_{t,i} \\
    &= \sum_{i \in \text{\mlstm}} \frac{\exp(z_{t,i})}{\sum_{j=1}^{E} \exp(z_{t,j})} \\
    &= \frac{\sum_{i \in \text{\mlstm}} \exp(\tilde{z}_{t,i} + \delta_{t,i})}{\sum_{j=1}^{E} \exp(z_{t,j})} \\
    &= \frac{\sum_{i \in \text{\mlstm}} \exp(\tilde{z}_{t,i} + \gamma d_t)}{\sum_{j=1}^{E} \exp(z_{t,j})} \\
    &= \frac{\exp(\gamma d_t) \sum_{i \in \text{\mlstm}} \exp(\tilde{z}_{t,i})}{\sum_{j=1}^{E} \exp(z_{t,j})}
    \label{eq:prob_mLSTM}
\end{align}

Next, let's express $P(\text{\slstm}|d_t)$:
\begin{align}
P(\text{\slstm}|d_t) &= \sum_{k \in \text{\slstm}} p_{t,k} \\
&= \sum_{k \in \text{\slstm}} \frac{\exp(z_{t,k})}{\sum_{j=1}^{E} \exp(z_{t,j})} \\
&= \frac{\sum_{k \in \text{\slstm}} \exp(\tilde{z}_{t,k} + \delta_{t,k})}{\sum_{j=1}^{E} \exp(z_{t,j})} \\
&= \frac{\sum_{k \in \text{\slstm}} \exp(\tilde{z}_{t,k} - \gamma d_t)}{\sum_{j=1}^{E} \exp(z_{t,j})} \\
&= \frac{\exp(-\gamma d_t) \sum_{k \in \text{\slstm}} \exp(\tilde{z}_{t,k})}{\sum_{j=1}^{E} \exp(z_{t,j})} \label{eq:prob_sLSTM}
\end{align}

Now, we compute the ratio using equations \eqref{eq:prob_mLSTM} and \eqref{eq:prob_sLSTM}:
\begin{align}
    \frac{P(\text{\mlstm}|d_t)}{P(\text{\slstm}|d_t)} &= \frac{\frac{\exp(\gamma d_t) \sum_{i \in \text{\mlstm}} \exp(\tilde{z}_{t,i})}{\sum_{j=1}^{E} \exp(z_{t,j})}}{\frac{\exp(-\gamma d_t) \sum_{k \in \text{\slstm}} \exp(\tilde{z}_{t,k})}{\sum_{j=1}^{E} \exp(z_{t,j})}} \\
    &= \frac{\exp(\gamma d_t) \sum_{i \in \text{\mlstm}} \exp(\tilde{z}_{t,i})}{\exp(-\gamma d_t) \sum_{k \in \text{\slstm}} \exp(\tilde{z}_{t,k})} \\
    &= \exp(\gamma d_t - (-\gamma d_t)) \frac{\sum_{i \in \text{\mlstm}} \exp(\tilde{z}_{t,i})}{\sum_{k \in \text{\slstm}} \exp(\tilde{z}_{t,k})} \\
    &= \exp(2\gamma d_t) \frac{\sum_{i \in \text{\mlstm}} \exp(\tilde{z}_{t,i})}{\sum_{k \in \text{\slstm}} \exp(\tilde{z}_{t,k})} \label{eq:ratio_exact}
\end{align}

\subsection*{Approximation and Final Result}

The expression \eqref{eq:ratio_exact} gives the exact ratio. If we assume that the router's preference based on the \textit{raw} logits $\tilde{z}_t$ is roughly balanced between the two groups of experts, meaning:

\begin{equation}
\sum_{i \in \text{\mlstm}} \exp(\tilde{z}_{t,i}) \approx \sum_{k \in \text{\slstm}} \exp(\tilde{z}_{t,k}) \label{eq:approximation}
\end{equation}

Then, the fraction in equation \eqref{eq:ratio_exact} is approximately equal to $1$. Under this assumption, the ratio simplifies to:
\begin{equation}
\frac{P(\text{\mlstm}|d_t)}{P(\text{\slstm}|d_t)} \approx \exp(2\gamma d_t)
\end{equation}

This final exponential relationship demonstrates that, under the assumption of balanced raw logits \eqref{eq:approximation}, the relative probability of routing to an mLSTM expert compared to an sLSTM expert grows exponentially with the token difficulty $d_t$, scaled by $2\gamma$.

The ratio of probabilities for routing to mLSTM versus sLSTM assuming similar base logits can be approximated as:

\begin{align}
\frac{P(mLSTM|d_t)}{P(sLSTM|d_t)} &= \frac{\exp(\tilde{z_t}^{mLSTM} + \delta_t)}{\exp(\tilde{z_t}^{sLSTM} - \delta_t)} \\
    &= \exp(\tilde{z_t}^{mLSTM} + \delta_t - \tilde{z_t}^{sLSTM} + \delta_t) \\
    &\approx \exp(2\delta_t) = \exp(2\gamma d_t)
\end{align}

This exponential relationship means that as token difficulty increases, the likelihood of routing to \mlstm \, experts increases exponentially, ensuring that difficult tokens receive the computational resources they need for accurate processing.

\section{Experimental Setup and Results}
\label{sec:experiments}

This section details the experimental configuration and presents results demonstrating the effectiveness and efficiency of our proposed MoxE architecture. We trained MoxE alongside Transformer (Gemma 2, Qwen1.5-MoE) and xLSTM baselines, configured as detailed in Table~\ref{tab:train_baselines_configuration}. Training utilized over 5 million tokens from the annotated portion of Fineweb-Edu \cite{lozhkov2024fineweb-edu} for one epoch. Model performance was validated using the unannotated version of Fineweb-Edu with a context length of 256 tokens.

\begin{figure}[H]
    \centering
    \begin{subfigure}[b]{0.48\textwidth}
        \includegraphics[width=\linewidth]{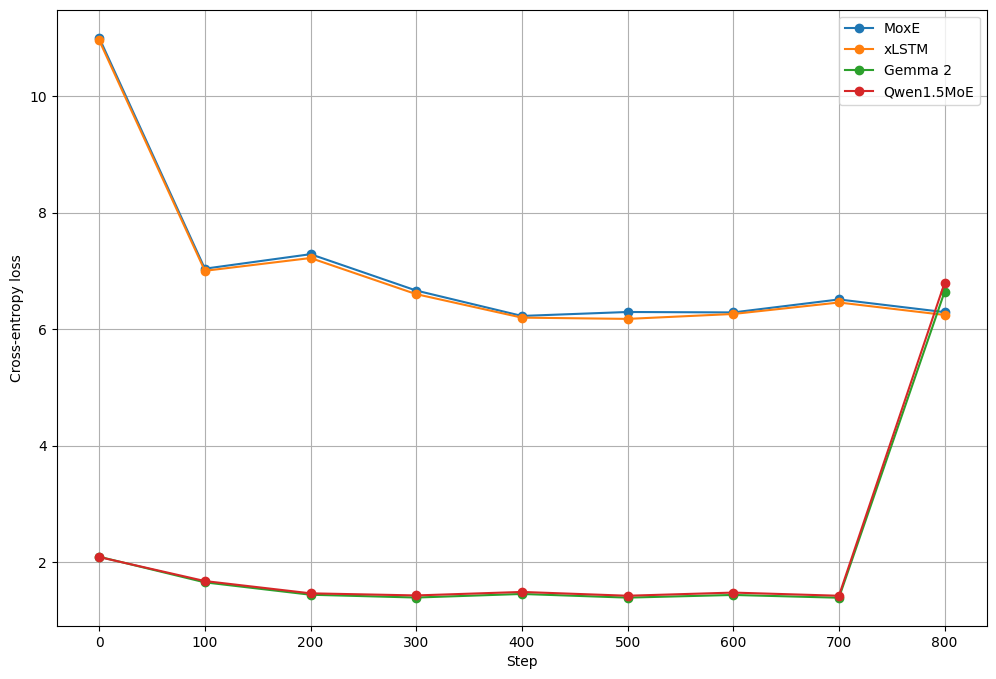}
        \caption{Cross-entropy loss during training}
        \label{fig:baseline_train_ce_loss}
    \end{subfigure}
    \hfill 
    \begin{subfigure}[b]{0.48\textwidth}
        \includegraphics[width=\linewidth]{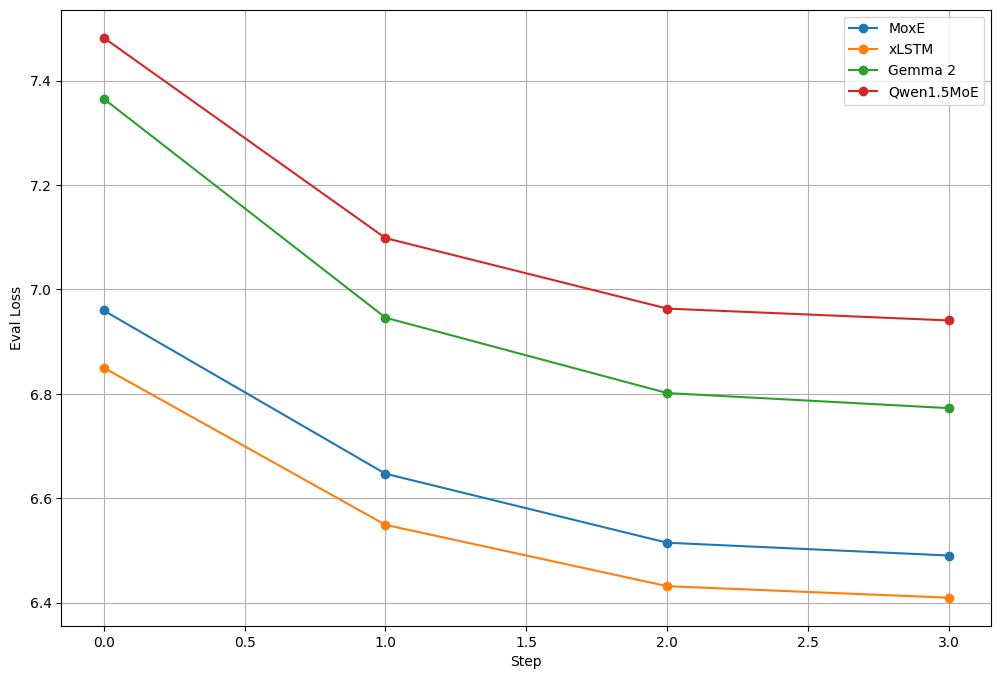}
        \caption{Evaluation loss on Fineweb-Edu}
        \label{fig:baseline_fineweb_eval_loss}
    \end{subfigure}

    \vspace{1em}

    \begin{subfigure}[b]{0.48\textwidth}
        \includegraphics[width=\linewidth]{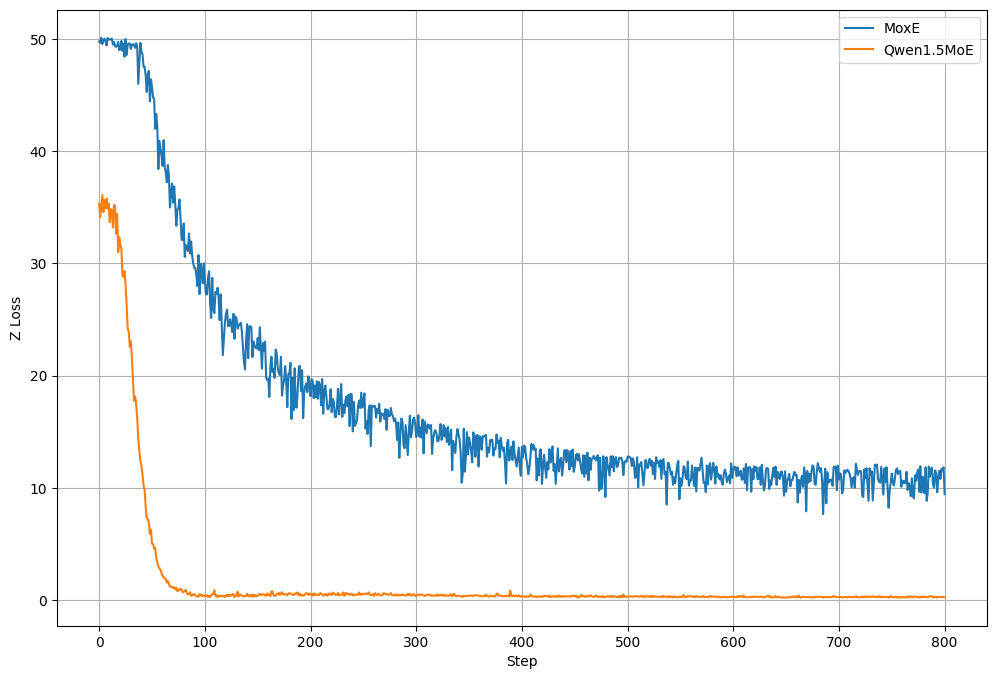}
        \caption{Router Z-loss during training}
        \label{fig:baseline_train_z_loss}
    \end{subfigure}
    \hfill 
    \begin{subfigure}[b]{0.48\textwidth}
        \includegraphics[width=\linewidth]{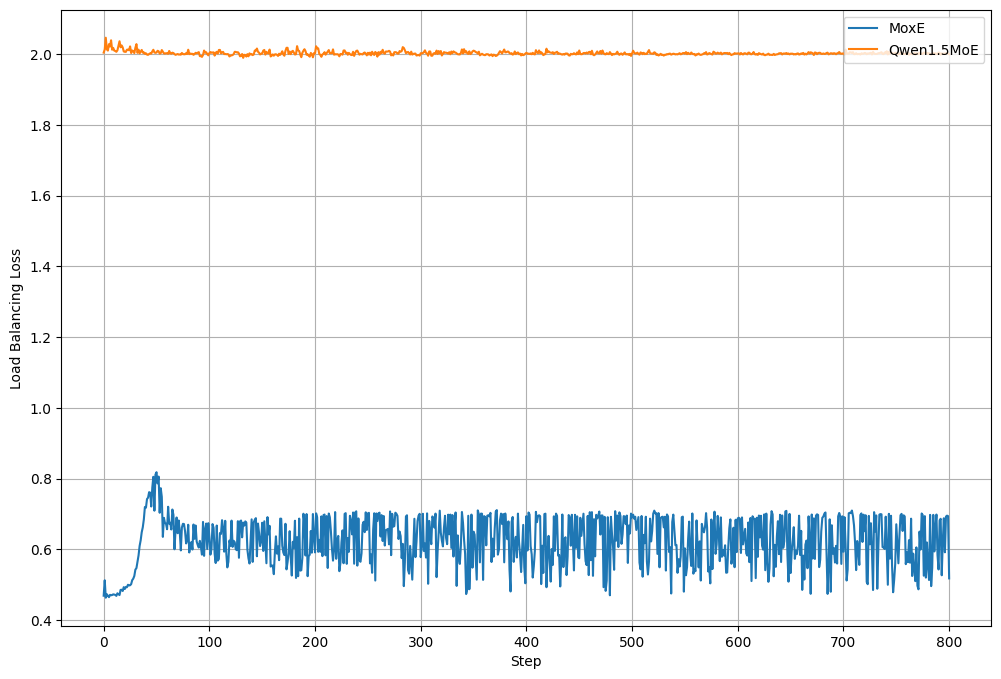}
        \caption{Load balancing loss during training}
        \label{fig:baseline_load_balancing_loss}
    \end{subfigure}
    
    \caption{Cross-entropy loss and evaluation of baseline models on Fineweb-Edu}
    \label{fig:baseline_training_metrics}
\end{figure}

\begin{figure}[H]
    \centering
    \includegraphics[width=0.8\linewidth]{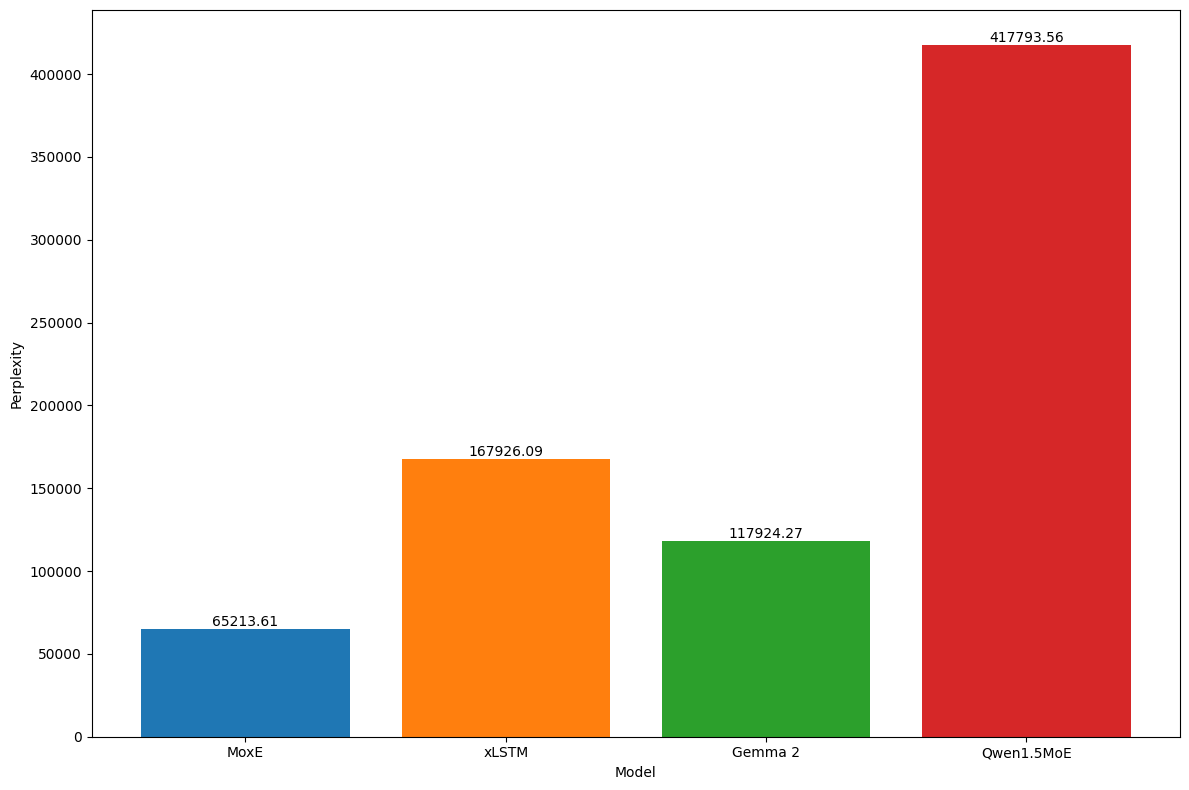}
    \caption{Baseline models' average perplexity on Lambada OpenAI}
    \label{fig:baseline_lambada_perplexity}
\end{figure}

On the Fineweb-Edu validation set (Figure~\ref{fig:baseline_fineweb_eval_loss}), MoxE achieves performance comparable to the xLSTM baseline, with both models yielding the lowest evaluation loss among the tested architectures. To evaluate generalization capabilities, we assessed perplexity on the Lambada OpenAI dataset \cite{paperno2016lambada} (Figure~\ref{fig:baseline_lambada_perplexity}). In this next-token prediction task, MoxE significantly outperforms both the Transformer and xLSTM baselines. Training dynamics, including cross-entropy loss (Figure~\ref{fig:baseline_train_ce_loss}), router Z-loss (Figure~\ref{fig:baseline_train_z_loss}), and load balancing loss (Figure~\ref{fig:baseline_load_balancing_loss}), are presented below.

\begin{table}[h]
    \centering
    \caption{Configuration of the baseline models used when evaluating MoxE}
    \begin{adjustbox}{width=\linewidth}
        \begin{tabular}{|c|c|c|c|c|c|c|c|}
            \hline
            Model & Embedding & Num Layers  & Num Heads & Num Experts & TopK & Learning Rate  & Parameters  \\
            \hline
            Gemma 2 & 640 & 15 & 8 & - & - & $3 \cdot 10^{-5}$ & 356M \\
            Qwen1.5-MoE & 640 & 10 & 16 & 6 & 2 & $3 \cdot 10^{-5}$ & 350M \\
            xLSTM[5:1]\footnotemark[1] & 1024 & 36 & 8 & - & - & $5 \cdot 10^{-5}$  & 338M \\
            MoxE & 640 & 10 & 4 & 8 & 2 & $5 \cdot 10^{-5}$ & 340M \\
            \hline
        \end{tabular}
    \end{adjustbox}
    \label{tab:train_baselines_configuration}
\end{table}

\footnotetext[1]{[5:1] refers to the ratio of mLSTM to sLSTM blocks used. After each 1 sLSTM block, 5 mLSTM blocks are stacked.}

\subsection{Ablation Studies}
To rigorously evaluate the contribution of each component within our proposed MoxE architecture, we conducted a series of ablation studies on the Lambada OpenAI dataset. We measured the impact on perplexity (PPL), with lower values indicating better performance. The key results are summarized in Table~\ref{tab:ablation_studies} and discussed below.

\begin{table}[htbp]
    \centering
    \caption{Ablation study results on the Lambada OpenAI dataset. Perplexity (PPL) values are reported, along with the percentage increase relative to the full MoxE baseline. Lower PPL is better.}
    \label{tab:ablation_studies}
    \begin{tabularx}{\textwidth}{|X|r|r|}
        \hline 
        Configuration & Lambada PPL & \% Increase vs MoxE \\
        \hline 
        \textbf{Full \moxe (Baseline)} & \textbf{65,213.61} & - \\
        \hline 
        \multicolumn{3}{|l|}{\textit{Ablating Core Components:}} \\
        \hline 
        No Entropy Bias with $\gamma = 0$ (\nobias) & 348,908.14 & +435.02\% \\
        No Group-Wise Loss & 213,974.42 & +228.11\% \\
        Replace xLSTM with \ffn{} and no entropy bias (\xlstmmoe) & 1,412,846.01 & +2066.48\% \\
        \hline 
        \multicolumn{3}{|l|}{\textit{Homogeneous Experts and no entropy bias:}} \\
        \hline 
        \mlstm{}-only Experts (\mlstmmoe) & 85,963.85 & +31.82\% \\
        \slstm{}-only Experts (\slstmmoe) & 191,161.87 & +193.14\% \\
        \hline 
    \end{tabularx}
\end{table}

\begin{figure}[ht]
    \centering
    \begin{subfigure}[b]{0.48\textwidth}
        \includegraphics[width=\linewidth]{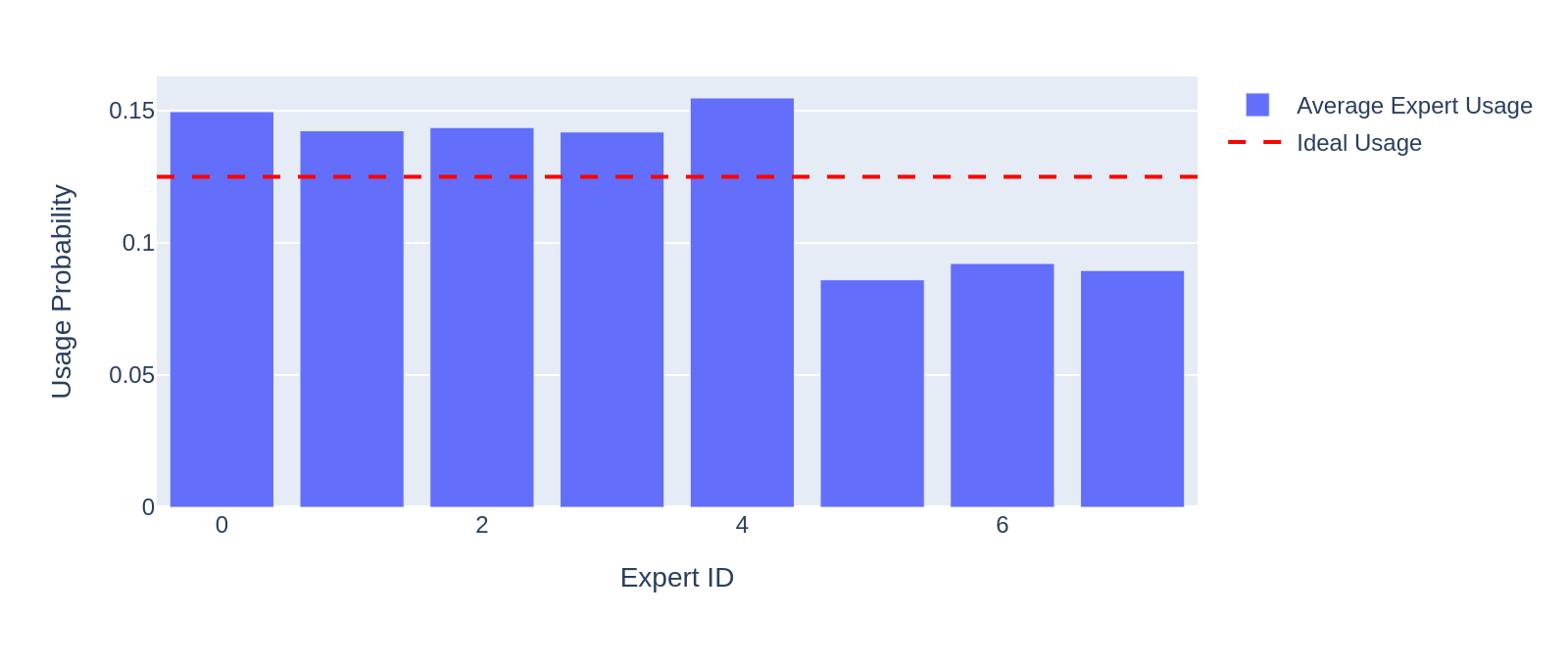}
        \caption{Layer 0 with group-wise loss}
        \label{fig:ablations_moxe_with_group_wise_l0_expert_usage}
    \end{subfigure}
    \hfill 
    \begin{subfigure}[b]{0.48\textwidth}
        \includegraphics[width=\linewidth]{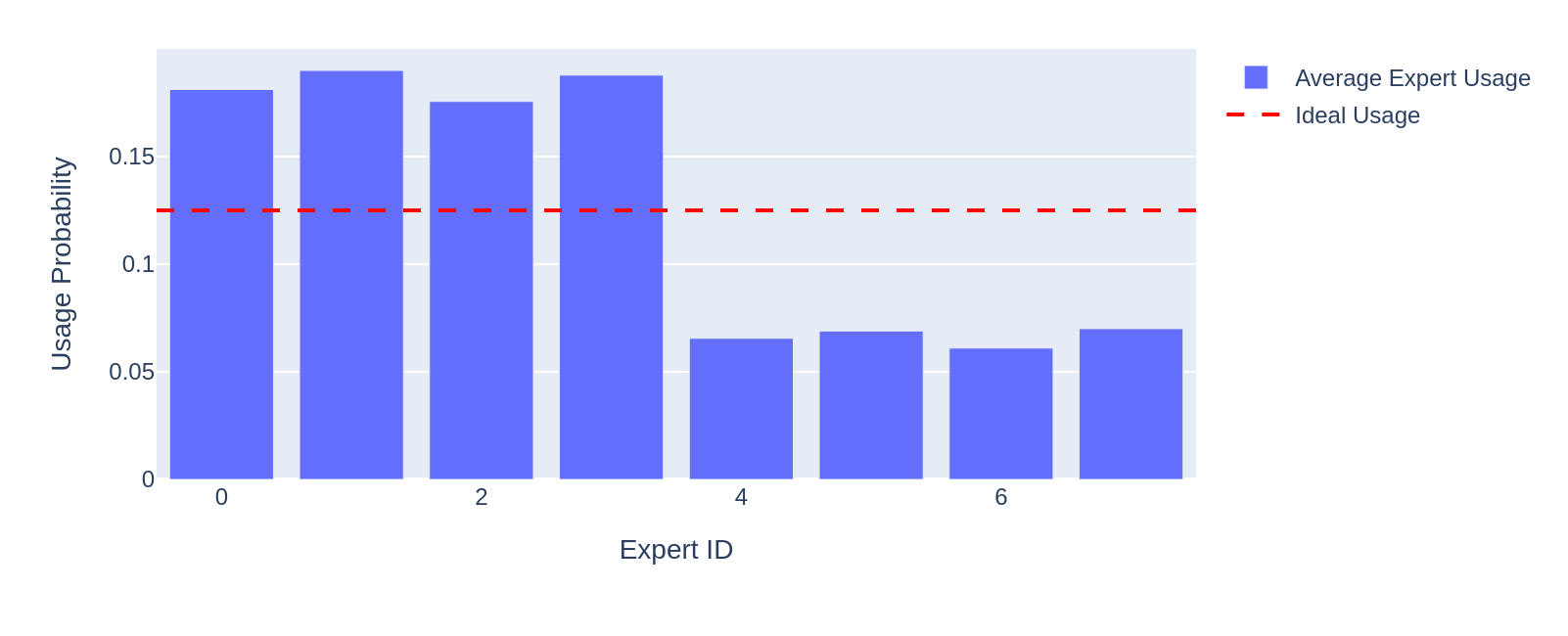}
        \caption{Layer 0 without group-wise loss}
        \label{fig:ablations_moxe_without_group_wise_l0_expert_usage}
    \end{subfigure}

    \vspace{1em}

    \begin{subfigure}[b]{0.48\textwidth}
        \includegraphics[width=\linewidth]{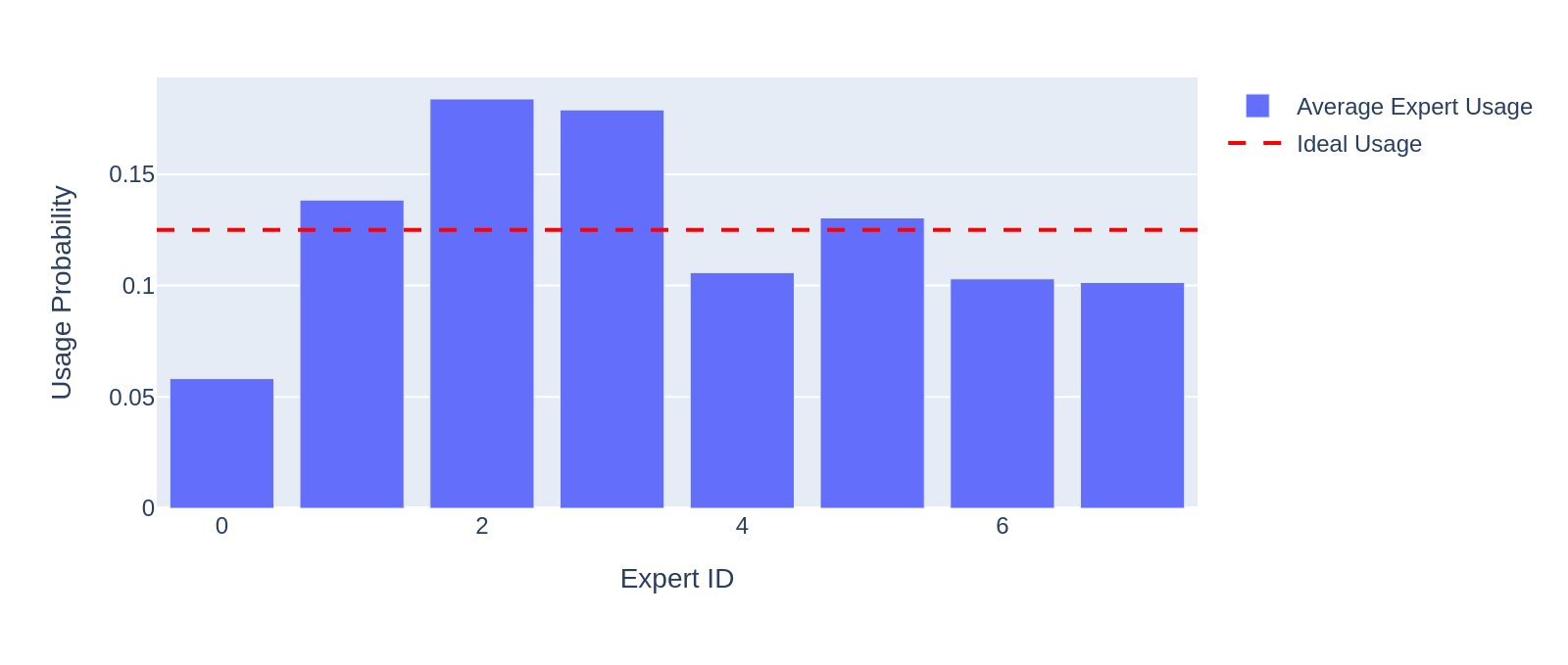}
        \caption{Layer 6 with group-wise loss}
        \label{fig:ablations_moxe_with_group_wise_l5_expert_usage}
    \end{subfigure}
    \hfill 
    \begin{subfigure}[b]{0.48\textwidth}
        \includegraphics[width=\linewidth]{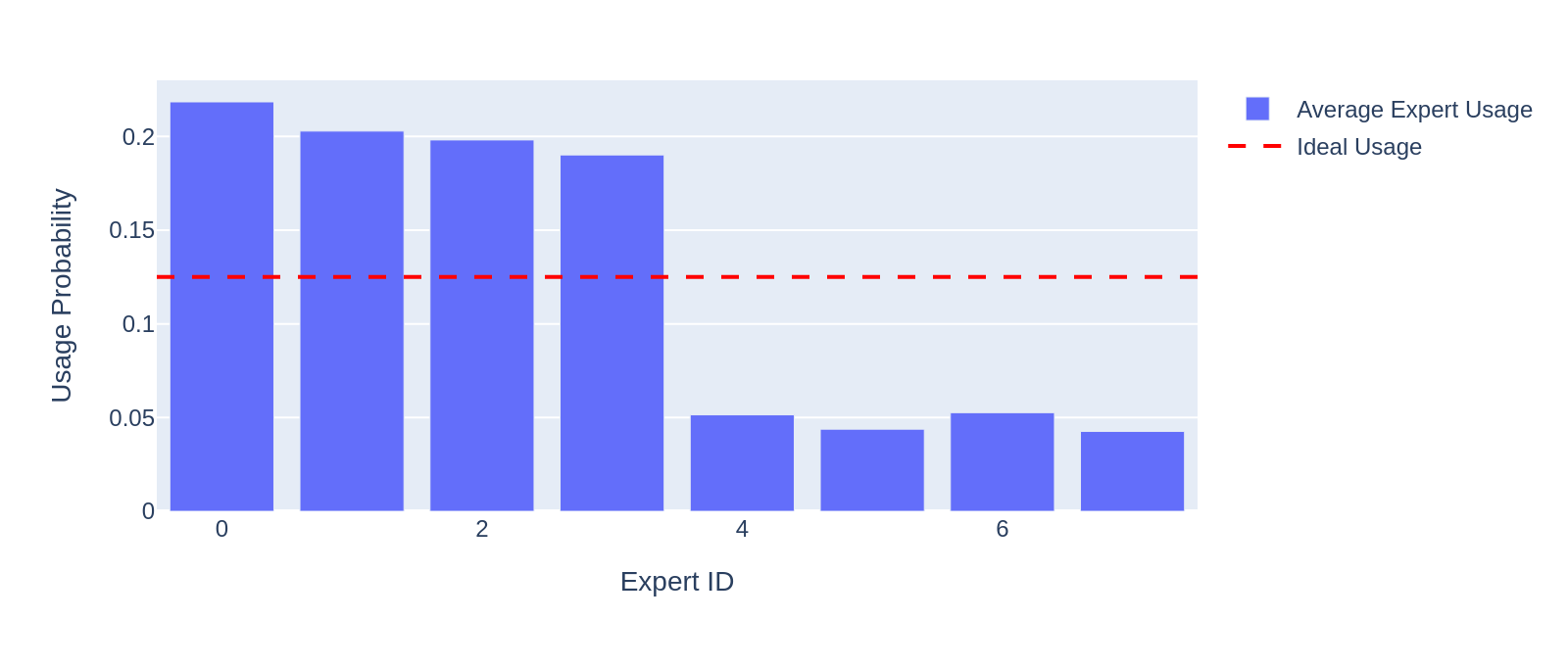}
        \caption{Layer 6 without group-wise loss}
        \label{fig:ablations_moxe_without_group_wise_l5_expert_usage}
    \end{subfigure}
    
    \vspace{1em}

    \begin{subfigure}[b]{0.48\textwidth}
        \includegraphics[width=\linewidth]{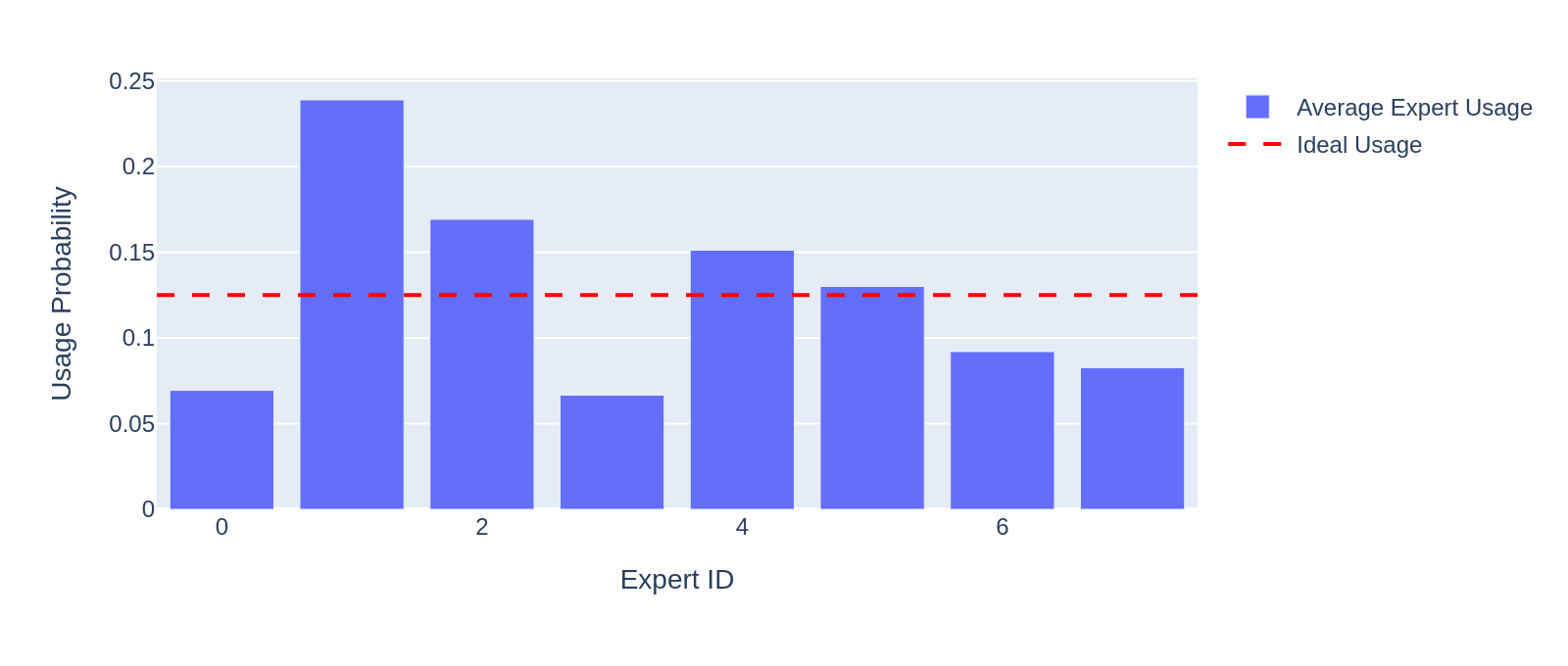}
        \caption{Layer 9 with group-wise loss}
        \label{fig:ablations_moxe_with_group_wise_l8_expert_usage}
    \end{subfigure}
    \hfill 
    \begin{subfigure}[b]{0.48\textwidth}
        \includegraphics[width=\linewidth]{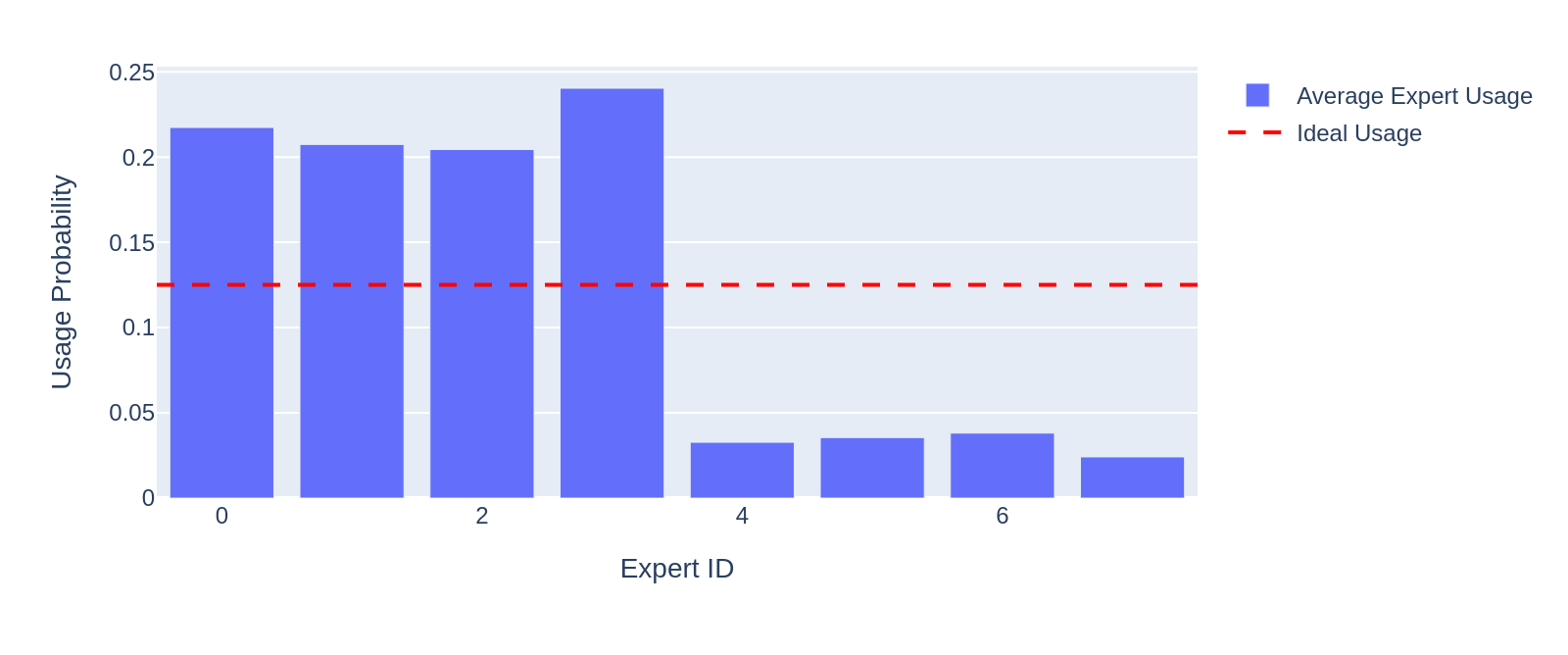}
        \caption{Layer 9 without group-wise loss}
        \label{fig:ablations_moxe_without_group_wise_l8_expert_usage}
    \end{subfigure}
    \caption{Average expert usage at layer 1, 6 and 9 of a MoxE trained with group-wise loss (left) and without group-wise loss (right) with the first four experts (Expert 0 to 3) being all \mlstm units and others \slstm units.}
    \label{fig:ablations_avg_expert_usage_group_loss_case}
\end{figure}

Our full MoxE model serves as the baseline, achieving a perplexity of 65,213.61. The ablation experiments reveal the following:

\begin{itemize}
    \item \textbf{Importance of Entropy-Based Routing Bias:} Removing the difficulty-aware routing mechanism ($\gamma = 0$, configuration \nobias{}) results in a substantial 435.02\% increase in perplexity to 348,908.14. This drastically poorer performance underscores the critical role of dynamically routing tokens based on their predicted difficulty ($\S$\ref{sec:analysis}).

    \item \textbf{Value of xLSTM Experts:} Replacing the specialized \mlstm{} and \slstm{} blocks with standard Feed-Forward Network (\ffn{}) experts (configuration \xlstmmoe{}) leads to the most significant performance degradation, increasing perplexity by a factor of over 20 (+2066.48\%) to 1,412,846.01. This clearly demonstrates the advantage of the enhanced memory and computational capabilities inherent in the xLSTM blocks for this task compared to simpler FFNs.

    \item \textbf{Necessity of Group-Wise Balancing:} Omitting the group-wise auxiliary loss ($\mathcal{L}_{group}$, configuration \nogrouploss{}) increases perplexity by 228.11\% to 213,974.42. Beyond the perplexity increase, this ablation leads to highly unbalanced expert utilization during training (as potentially shown in Figure~\ref{fig:ablations_avg_expert_usage_group_loss_case}). Without this loss, the routing mechanism, influenced by the modulation bias defined in Eq.~\eqref{eq:modulation_bias}, tends to disproportionately favor the \mlstm{} experts, neglecting the \slstm{} group and hindering overall model effectiveness ($\S$\ref{sec:losses}).

    \item \textbf{Benefit of Heterogeneous Experts:} We compared the baseline MoxE to variants using only one type of xLSTM expert. Using only \mlstm{} experts (\mlstmmoe{}) increased perplexity by 31.82\% (to 85,963.85), while using only \slstm{} experts (\slstmmoe{}) resulted in a larger 193.14\% increase (to 191,161.87). Both homogeneous configurations perform worse than the mixed-expert MoxE baseline strongly suggests that the heterogeneity is beneficial. It allows the model to leverage the potentially distinct strengths of \mlstm{} and \slstm{} architectures, guided by the difficulty-based routing, achieving better overall performance than relying on a single expert type.
\end{itemize}

In summary, these ablation studies, with results quantified in Table~\ref{tab:ablation_studies}, consistently highlight that each evaluated component, the heterogeneous \mlstm{}/\slstm{} experts, the entropy-aware routing bias, and the group-wise balancing loss make a significant and positive contribution to the performance of the MoxE architecture. The synergy between these components appears crucial for achieving the reported results.

\begin{figure}[H]
    \centering
    \begin{subfigure}[b]{0.48\textwidth}
        \includegraphics[width=\linewidth]{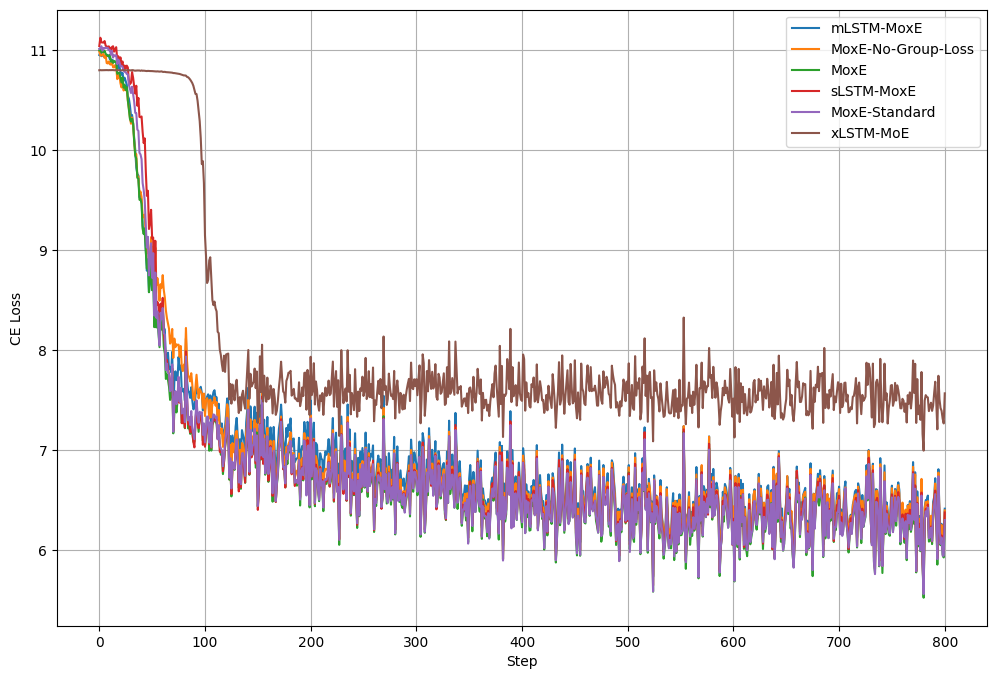}
        \caption{Cross-entropy loss during training}
        \label{fig:ablations_train_ce_loss}
    \end{subfigure}
    \hfill 
    \begin{subfigure}[b]{0.48\textwidth}
        \includegraphics[width=\linewidth]{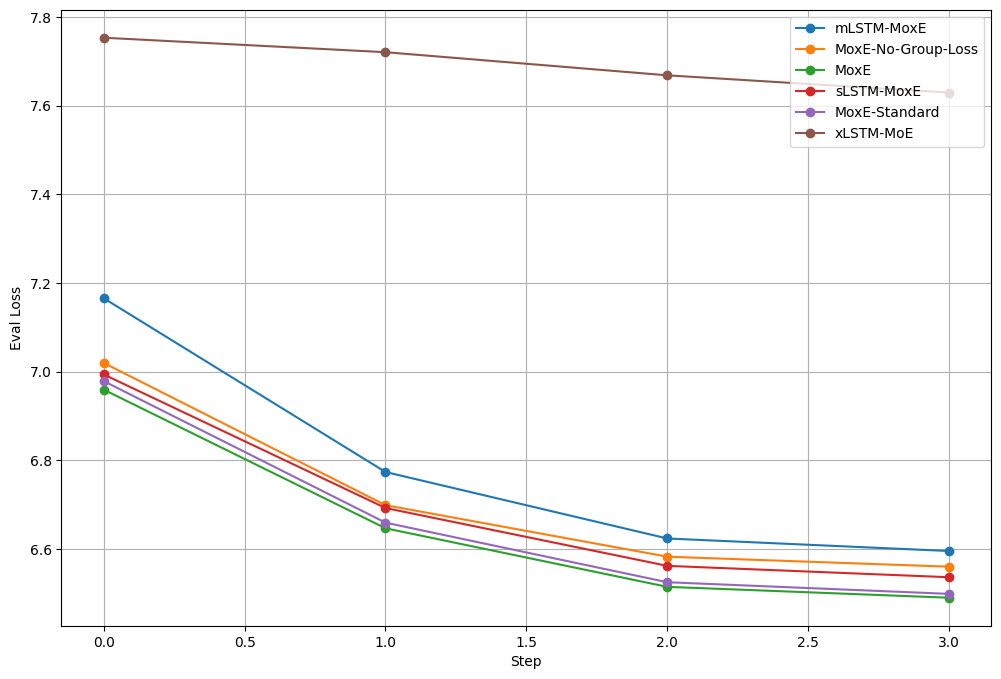}
        \caption{Evaluation loss on Fineweb-Edu}
        \label{fig:ablations_fineweb_eval_loss}
    \end{subfigure}
    \caption{Training loss and evaluation on Fineweb-Edu after our conducted ablation studies}
    \label{fig:ablations_training_metrics}
\end{figure}

\begin{figure}[H]
    \centering
    \begin{subfigure}[b]{0.48\textwidth}
        \includegraphics[width=\linewidth]{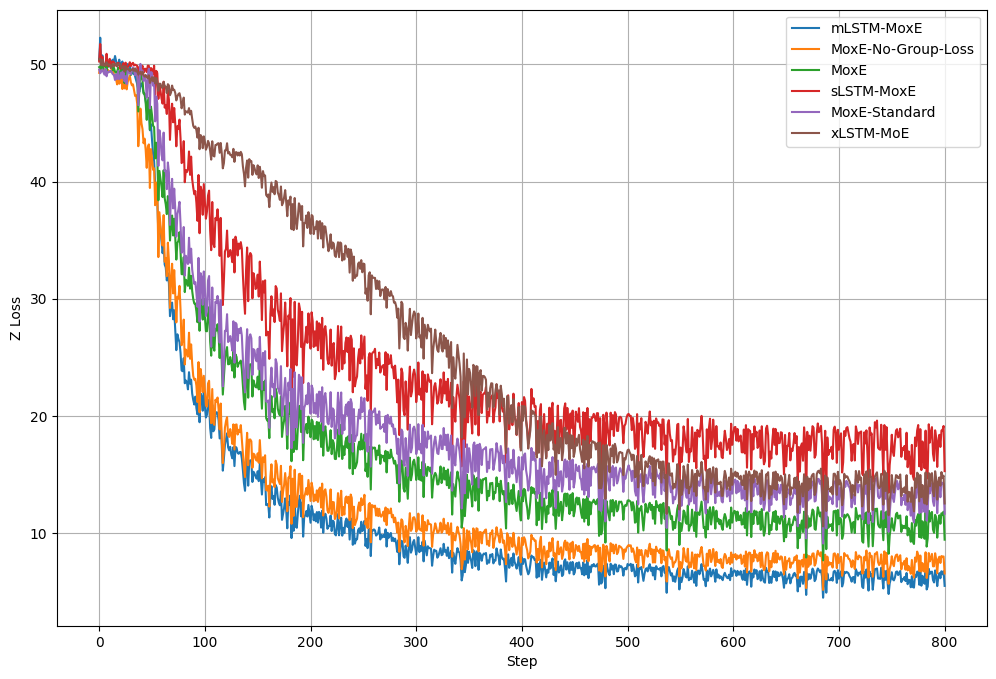}
        \caption{Router Z-loss during training}
        \label{fig:ablations_train_z_loss}
    \end{subfigure}
    \hfill 
    \begin{subfigure}[b]{0.48\textwidth}
        \includegraphics[width=\linewidth]{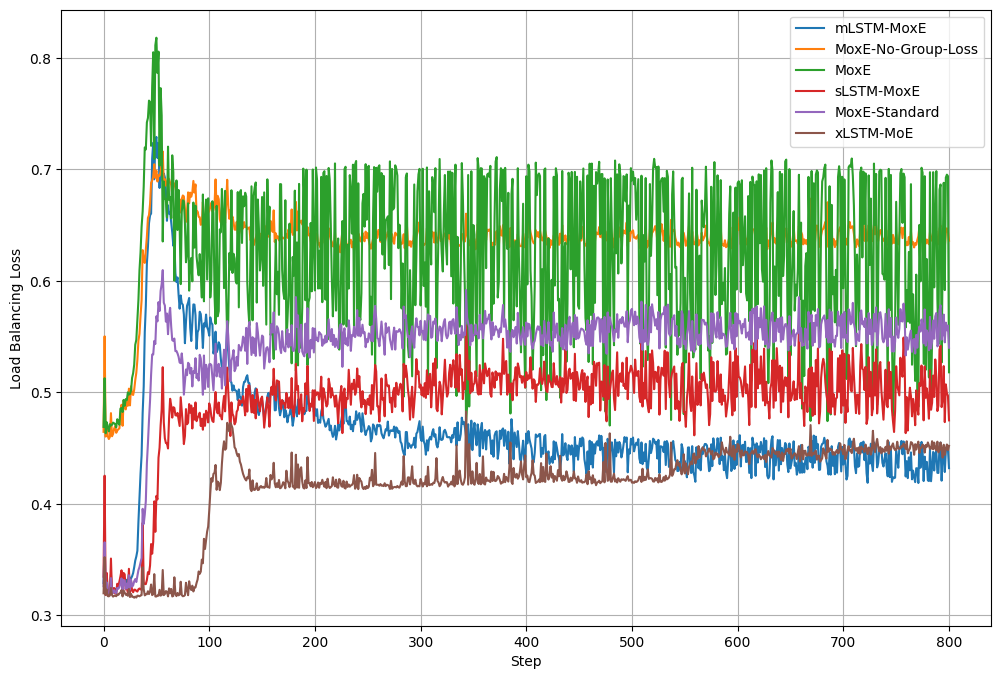}
        \caption{Load balancing loss during training}
        \label{fig:ablations_load_balancing_loss}
    \end{subfigure}

    \vspace{1em}

    \begin{subfigure}[b]{0.48\textwidth}
        \includegraphics[width=\linewidth]{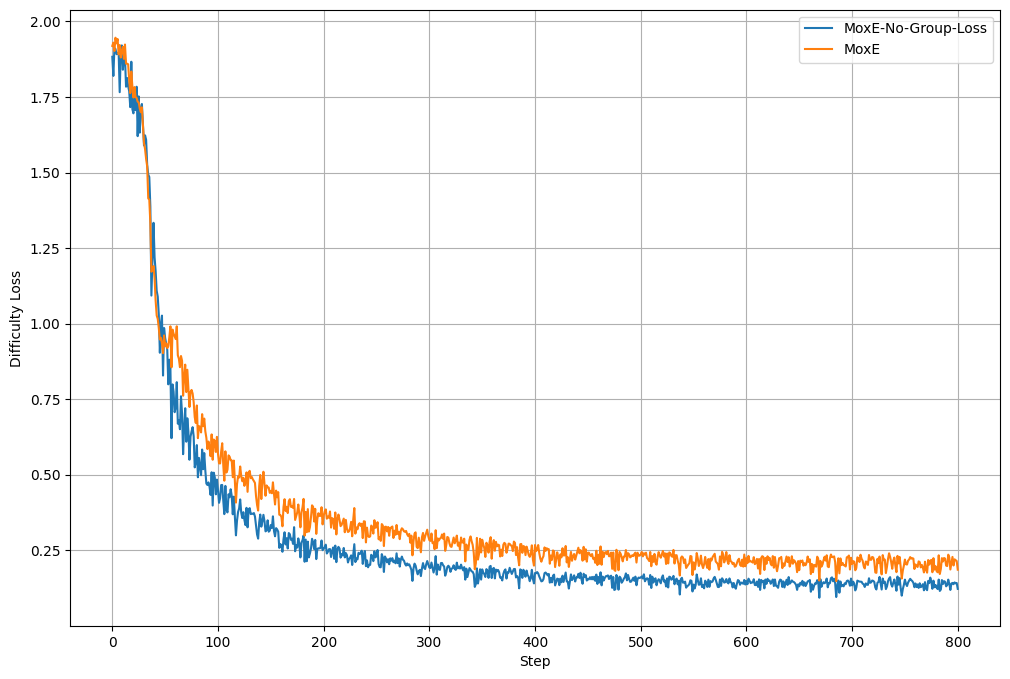}
        \caption{Difficulty loss during training}
        \label{fig:ablations_d_loss}
    \end{subfigure}
    
    \caption{Training Z-loss, auxiliary load balancing and difficulty losses on Fineweb-Edu for ablation studies}
    \label{fig:ablations_training_aux_losses}
\end{figure}

\begin{figure}[h]
    \centering
    \includegraphics[width=0.8\linewidth]{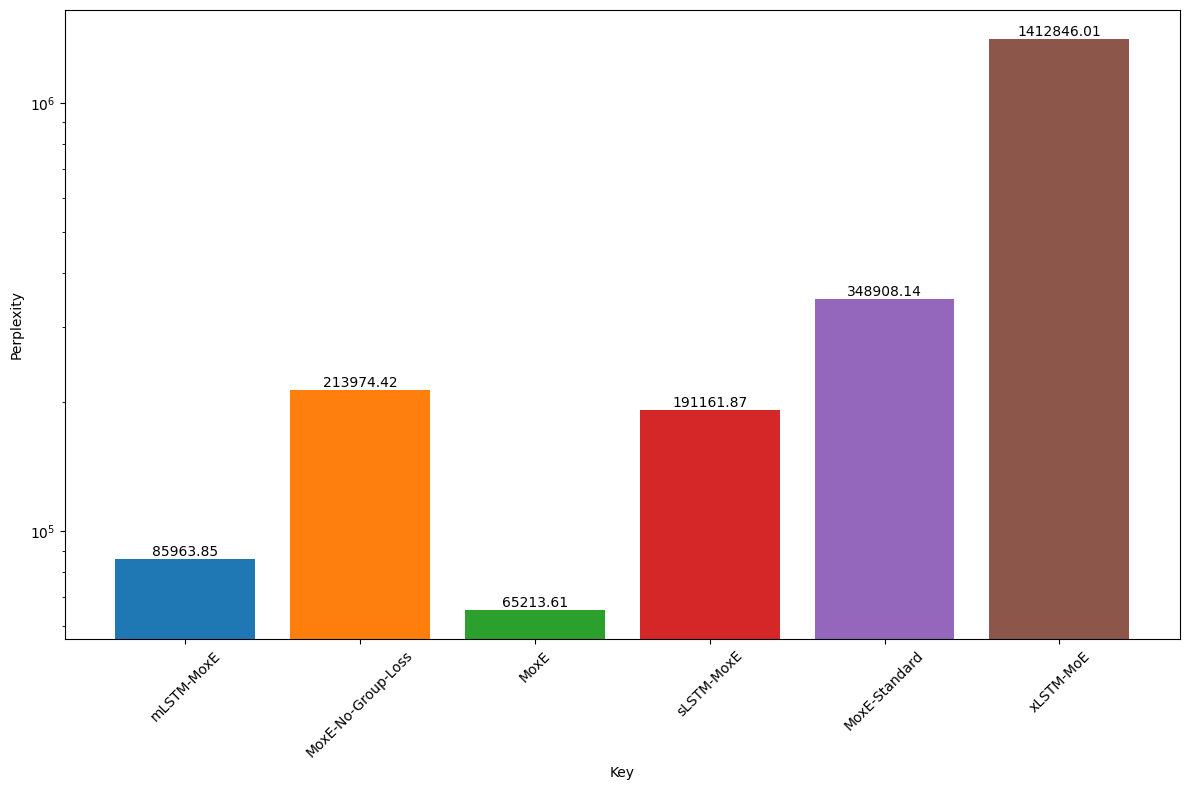}
    \caption{Log-scaled average perplexity on Lambada OpenAI of various MoxE models to conduct ablation studies}
    \label{fig:ablations_lambada_perplexity}
\end{figure}

\section{Related Work}
\label{sec:related_work}

\subsection{Recurrent Neural Networks and Extensions}
Traditional recurrent neural networks, including LSTMs \cite{hochreiter1997lstm}, have played a crucial role in sequence modeling tasks. Recent work on extending these architectures includes xLSTM \cite{beck2025xlstm}, which introduces novel memory structures and computational units to address the limitations of standard recurrent models. Our work builds upon xLSTM by incorporating it into a sparse MoE framework.

\subsection{Mixture of Experts Models}
The Mixture of Experts approach has a long history in machine learning \cite{jacobs1991adaptive}, with recent resurgence in the context of large language models. Notable works include Switch Transformers \cite{fedus2022switch}, which demonstrated the effectiveness of sparse gating in scaling transformer models to trillions of parameters. GShard \cite{lepikhin2020gshardscalinggiantmodels} explored techniques for load balancing and efficient distributed training of MoE models. Our work differs from these approaches by focusing on recurrent experts rather than feedforward networks.

\subsection{Recent Advances in MoE Routing}
Recent work has focused on improving routing mechanisms in MoE models. GW-MoE \cite{wu2024gwmoeresolvinguncertaintymoe} addresses the uncertainty in MoE router modules during fine-tuning, introducing techniques to make routing more robust. Huang et al. \cite{huang2024hardertasksneedexperts} showed that harder tasks benefit from more experts, introducing dynamic routing based on task difficulty. Our approach is inspired by these insights, incorporating difficulty-based routing into our architecture.

\subsection{Recurrent MoE Models}
There has been growing interest in combining recurrent architectures with the MoE framework. MoE-Mamba \cite{pióro2024moemambaefficientselectivestate} integrates the Mamba architecture with sparse MoE, demonstrating efficiency gains. Jamba \cite{lieber2024jambahybridtransformermambalanguage} explores a hybrid transformer-Mamba approach within the MoE framework. Our work differs from these approaches by focusing specifically on xLSTM as the base architecture and introducing entropy-aware routing tailored to the unique properties of mLSTM and sLSTM experts.

\section{Conclusion and Future Work}
\label{sec:conclusion}

In this paper, we introduced MoxE, a novel architecture that combines the efficiency of xLSTM with the scalability of sparse MoE models. By leveraging entropy-aware routing to direct tokens to specialized experts based on their difficulty, our approach achieves both computational efficiency and state-of-the-art performance on language modeling tasks.

The key contributions of our work include:
\begin{itemize}
    \item A fully recurrent MoE architecture that leverages the unique properties of mLSTM and sLSTM computational units.
    \item An entropy-aware routing mechanism that dynamically allocates computational resources based on token difficulty.
    \item A set of auxiliary losses that ensure balanced and effective training of the MoxE model.
    \item Comprehensive empirical evaluation demonstrating the efficiency and effectiveness of our approach.
\end{itemize}

Our results suggest that recurrent architectures, when combined with appropriate sparsity techniques, can be competitive with or superior to attention-based models in terms of both performance and efficiency. This challenges the dominant paradigm in NLP, which has been heavily focused on attention-based architectures.

Future work could explore several promising directions:
\begin{itemize}
    \item Scaling MoxE to even larger model sizes and investigating the scaling properties of recurrent MoE models.
    \item Adapting MoxE for specific downstream tasks beyond language modeling.
    \item Exploring more sophisticated routing mechanisms that incorporate additional signals beyond token entropy.
    \item Investigating the potential of MoxE for efficient fine-tuning and adaptation to new domains.
\end{itemize}

Overall, MoxE represents a significant step towards more efficient and effective language models, offering a compelling alternative to the dominant attention-based architectures in the NLP space.

\bibliographystyle{spbasic_updated}
\bibliography{main}

\begin{thebibliography}{13}
\providecommand{\natexlab}[1]{#1}
\providecommand{\url}[1]{{#1}}
\providecommand{\urlprefix}{URL }
\expandafter\ifx\csname urlstyle\endcsname\relax
  \providecommand{\doi}[1]{DOI~\discretionary{}{}{}#1}\else
  \providecommand{\doi}{DOI~\discretionary{}{}{}\begingroup \urlstyle{rm}\Url}\fi
\providecommand{\eprint}[2][]{\url{#2}}

\bibitem[{Beck et~al.(2025)Beck, P{\"o}ppel, Spanring, Auer, Prudnikova, Kopp, Klambauer, Brandstetter, and Hochreiter}]{beck2025xlstm}
Beck M, P{\"o}ppel K, Spanring M, Auer A, Prudnikova O, Kopp M, Klambauer G, Brandstetter J, Hochreiter S (2025) xlstm: Extended long short-term memory. Advances in Neural Information Processing Systems 37:107,547--107,603

\bibitem[{Fedus et~al.(2022)Fedus, Zoph, and Shazeer}]{fedus2022switch}
Fedus W, Zoph B, Shazeer N (2022) Switch transformers: Scaling to trillion parameter models with simple and efficient sparsity. Journal of Machine Learning Research 23(120):1--39

\bibitem[{Hochreiter and Schmidhuber(1997)}]{hochreiter1997lstm}
Hochreiter S, Schmidhuber J (1997) Long short-term memory. Neural Comput 9(8):1735–1780, \doi{10.1162/neco.1997.9.8.1735}

\bibitem[{Huang et~al.(2024)Huang, An, Zhuang, Tao, Zhang, Jin, Xu, Chen, Huang, and Feng}]{huang2024hardertasksneedexperts}
Huang Q, An Z, Zhuang N, Tao M, Zhang C, Jin Y, Xu K, Chen L, Huang S, Feng Y (2024) Harder task needs more experts: Dynamic routing in moe models. In: Proceedings of the 62nd Annual Meeting of the Association for Computational Linguistics (Volume 1: Long Papers), pp 12,883--12,895

\bibitem[{Jacobs et~al.(1991)Jacobs, Jordan, Nowlan, and Hinton}]{jacobs1991adaptive}
Jacobs R, Jordan M, Nowlan S, Hinton G (1991) Adaptive mixtures of local experts. Neural Computation 3:79--87, \doi{10.1162/neco.1991.3.1.79}

\bibitem[{Lepikhin et~al.(2020)Lepikhin, Lee, Xu, Chen, Firat, Huang, Krikun, Shazeer, and Chen}]{lepikhin2020gshardscalinggiantmodels}
Lepikhin D, Lee H, Xu Y, Chen D, Firat O, Huang Y, Krikun M, Shazeer N, Chen Z (2020) Gshard: Scaling giant models with conditional computation and automatic sharding

\bibitem[{Lieber et~al.(2024)Lieber, Lenz, Bata, Cohen, Osin, Dalmedigos, Safahi, Meirom, Belinkov, Shalev-Shwartz, Abend, Alon, Asida, Bergman, Glozman, Gokhman, Manevich, Ratner, Rozen, Shwartz, Zusman, and Shoham}]{lieber2024jambahybridtransformermambalanguage}
Lieber O, Lenz B, Bata H, Cohen G, Osin J, Dalmedigos I, Safahi E, Meirom S, Belinkov Y, Shalev-Shwartz S, Abend O, Alon R, Asida T, Bergman A, Glozman R, Gokhman M, Manevich A, Ratner N, Rozen N, Shwartz E, Zusman M, Shoham Y (2024) Jamba: A hybrid transformer-mamba language model

\bibitem[{Lozhkov et~al.(2024)Lozhkov, Ben~Allal, von Werra, and Wolf}]{lozhkov2024fineweb-edu}
Lozhkov A, Ben~Allal L, von Werra L, Wolf T (2024) Fineweb-edu: the finest collection of educational content. \doi{10.57967/hf/2497}

\bibitem[{Paperno et~al.(2016)Paperno, Kruszewski, Lazaridou, Pham, Bernardi, Pezzelle, Baroni, Boleda, and Fernández}]{paperno2016lambada}
Paperno D, Kruszewski G, Lazaridou A, Pham QN, Bernardi R, Pezzelle S, Baroni M, Boleda G, Fernández R (2016) The lambada dataset. \doi{10.5281/zenodo.2630551}

\bibitem[{Pióro et~al.(2024)Pióro, Ciebiera, Król, Ludziejewski, Krutul, Krajewski, Antoniak, Miłoś, Cygan, and Jaszczur}]{pióro2024moemambaefficientselectivestate}
Pióro M, Ciebiera K, Król K, Ludziejewski J, Krutul M, Krajewski J, Antoniak S, Miłoś P, Cygan M, Jaszczur S (2024) Moe-mamba: Efficient selective state space models with mixture of experts

\bibitem[{Shazeer et~al.(2017)Shazeer, Mirhoseini, Maziarz, Davis, Le, Hinton, and Dean}]{shazeer2017outrageouslylargeneuralnetworks}
Shazeer N, Mirhoseini A, Maziarz K, Davis A, Le Q, Hinton G, Dean J (2017) Outrageously large neural networks: The sparsely-gated mixture-of-experts layer

\bibitem[{Waswani et~al.(2017)Waswani, Shazeer, Parmar, Uszkoreit, Jones, Gomez, Kaiser, and Polosukhin}]{waswani2017attention}
Waswani A, Shazeer N, Parmar N, Uszkoreit J, Jones L, Gomez A, Kaiser L, Polosukhin I (2017) Attention is all you need. In: NIPS

\bibitem[{Wu et~al.(2024)Wu, Qiu, Wang, Zhao, and Fu}]{wu2024gwmoeresolvinguncertaintymoe}
Wu H, Qiu Z, Wang Z, Zhao H, Fu J (2024) Gw-moe: Resolving uncertainty in moe router with global workspace theory

\end{thebibliography}
\end{document}